\documentclass[sigconf]{acmart}

\usepackage{booktabs}
\usepackage{graphicx}

\usepackage{algorithm}
\usepackage{algorithmic}
\usepackage{amsmath}

\setcopyright{rightsretained}

\acmConference[ICVGIP'24]{15th Indian Conference on Computer Vision, Graphics and Image Processing}{December 2024}{Bangalore, India}
\acmYear{2024}
\copyrightyear{2024}

\acmPrice{15.00}

\newcommand{\poze}{\texttt{poze~}}

\begin{document}
\title{Poze: Sports Technique Feedback under Data Constraints}
%\titlenote{Produces the permission block, and
 %copyright information}

\author{Agamdeep Singh$^1$, Sujit PB$^1$, and Mayank Vatsa$^2$}
\affiliation{%
  \institution{1- IISER Bhopal, 2-IIT Jodhpur}
  \streetaddress{}
  \city{}
  \state{}
  \country{}
  \postcode{}
}

%\author{Sujit PB}
%\affiliation{%
%  \institution{IISER Bhopal}
%  \streetaddress{}
%  \city{}
%  \state{}
%  \country{}
%  \postcode{}
%}

%\author{Mayank Vatsa}
%\affiliation{%
%  \institution{IIT Jodhpur}
%  \streetaddress{}
%  \city{}
%  \state{}
%  \country{}
%  \postcode{}
%}
\renewcommand{\shortauthors}{}

\begin{abstract}
Access to expert coaching is essential for developing technique in sports, yet economic barriers often place it out of reach for many enthusiasts. To bridge this gap, we introduce \textbf{Poze}—an innovative video processing framework that provides feedback on human motion, emulating the insights of a professional coach. \textit{Poze} combines pose estimation with sequence comparison and is optimized to function effectively with minimal data. \textit{Poze} surpasses state-of-the-art vision-language models in video question-answering frameworks, achieving 70\% and 196\% increase in accuracy over GPT-4V and LLaVA-v1.6-7b, respectively.

%Expert coaching is crucial for technique development in sports but often inaccessible to sports enthusiasts due to economic constraints. To address this, we present \poze, a novel video processing framework for generating coach-like feedback on human motion. It combines pose estimation with sequence comparison, and is designed to work with limited data. \poze outperforms state-of-the-art vision-language models in Video Question Answer frameworks, achieving 70\% and 196\% improvements in accuracy over GPT4-V and LLaVa-v1.6-7b, respectively.
\end{abstract}




\maketitle

%\maketitle

\section{Introduction}

In sports, access to quality coaching often defines the line between elite athletes and enthusiasts. This disparity is especially pronounced in developing nations like India. While deep learning-based applications offer potential solutions, they struggle to replicate the nuanced feedback of human coaches due to high data requirements and specialized hardware. Motion similarity techniques falter due to camera angle variations \cite{2d}, while motion embedding approaches like \cite{motionclip} lack stability for consistent feedback. Recent video question-answering (VQA) models show promise for video feedback but remain underexplored in sports analysis \cite{igvlm,videollava}. Furthermore, these methods are computationally and data expensive, requiring extensive data collection and specialized hardware for inference, which are often unavailable on portable devices.

India's passion for cricket presents an ideal case study; however, millions of aspiring players have limited access to professional coaching, hindering their athletic development. To address this, we introduce \textbf{\poze}, a representation learning framework designed to provide coach-like feedback using only smartphone video input. By combining pose estimation with a sequence comparison algorithm, \poze aims to overcome existing limitations while requiring minimal data.

\begin{figure}
    \centering
    \includegraphics[width=0.9\linewidth]{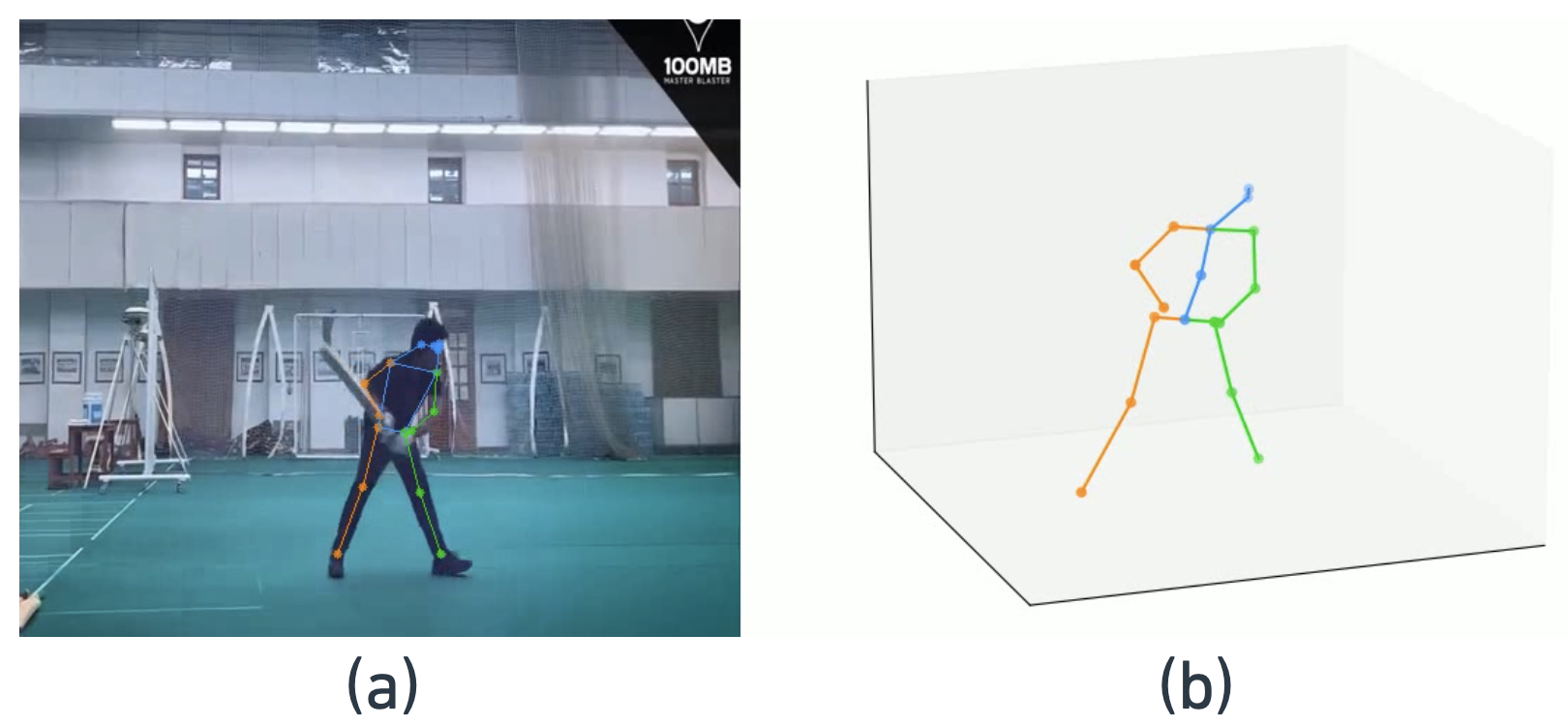}
    \caption{Pose estimation takes as input a video frame $f_i$ as shown in (a) and returns the 3D pose $p_i$ as in (b).}
    \label{fig:extraction}
\end{figure}
\section{Methodology}
The \poze framework consists of pose sequence extraction, pre-processing, building a generalised representation and then attribute classification as shown in Figure \ref{fig:pipe}. Let $V = \{f_1, \ldots, f_n\}$ be a video consisting of $n$ frames, where $f_i$ is the $i^{th}$ frame.
\subsection{Pose Sequence Extraction}
We extract 3D pose sequence $S = \{p_1, p_2, ..., p_n\}$ from $V$ using MotionBERT \cite{motionbert}, where $p_i \in \mathbb{R}^{J \times 3}$ represents a 3D skeleton of $J$ joints from frame $f_i$. MotionBERT tracks 17 human body joints as shown in Fig.\ref{fig:extraction} ($J=17$).

\subsection{Sequence Pre-Processing}
Pre-processing is done to account for variations in body proportions and video length. We perform pose normalization and temporal alignment between sequences using Dynamic Time Warping(DTW) \cite{dtw}.

\begin{figure*}
    \centering
    \includegraphics[width=0.9\linewidth]{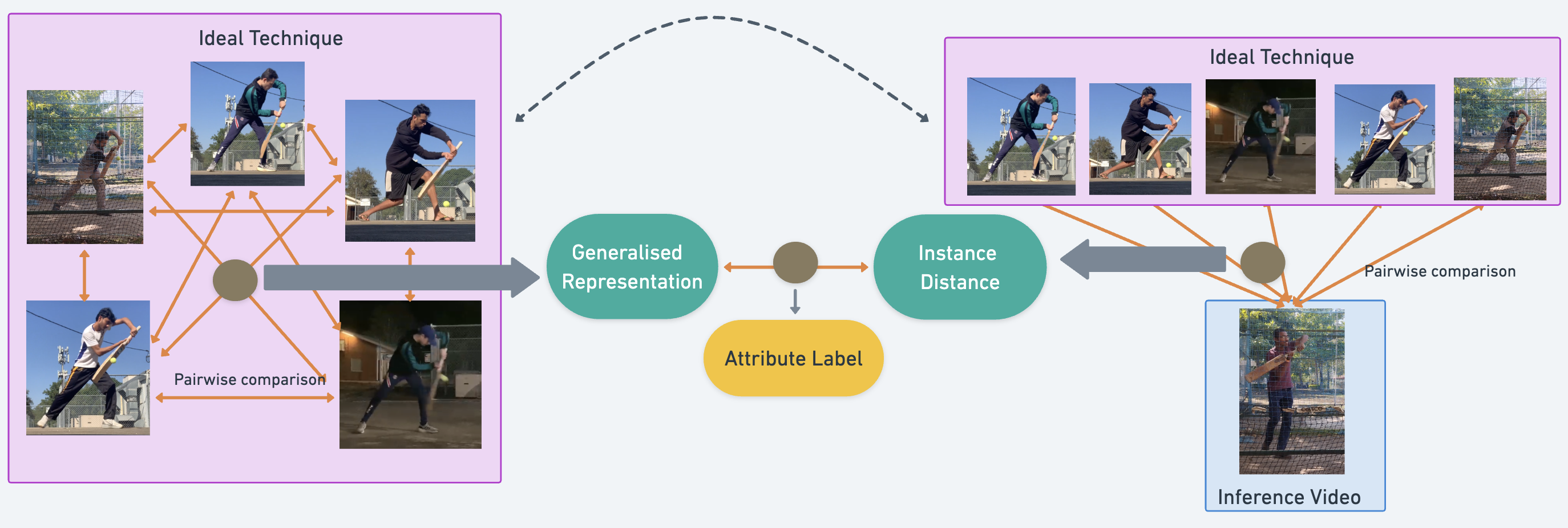}
    \caption{During inference, videos are compared with the ideal technique representation to get attribute labels.}
    \label{fig:pipe}
\end{figure*}

\subsection{Generalized Representation}

We build a generalized representation of the ideal technique execution, consisting of the mean error $\mu_j$, and the variance in error $\sigma_j$, for each joint $j$. Let $G_t = \{S_1,\ldots, S_m\}$ be a set of $m$ pose sequences extracted from videos where the technique is ideally executed by a coach or professional player. We calculate $\mu_{j}$ and $\sigma_{j}$ using Equation \eqref{eq:1} and \eqref{eq:2}, respectively.
\begin{eqnarray} \label{eq:1}
\mu_{j} &=& \frac{1}{m(m-1)} \sum_{i=1}^m \sum_{l=i+1}^m D^j_{i,l}\\
\sigma_{j} &=& \sqrt{\frac{1}{m(m-1)} \sum_{i=1}^m \sum_{l=i+1}^m (D^j_{i,l}- \mu_{j})^2}  \label{eq:2}
\end{eqnarray}
Here, $D^j_{i,l}$ is the time average distance between joint $j$ in the DTW-aligned sequences $S_i$ and $S_l$.

\subsection{Attribute Classification}

For a new video $V_{\text{new}}$ that needs feedback we extract pose sequence $S_{\text{new}}$. We then derive the instance distance $\bar{d}_j$, the average distance for joint $j$ in $S_{\text{new}}$ compared to all good examples in $G_t$. Then, we calculate a z-score as $z_{i,j} = \frac{\bar{d}_j - \mu_{j}}{\sigma_{j}}$.
The z-scores are binned for classification to qualitatively assess how well $V_{new}$ imitated the generalised representation of the ideal technique (e.g. very good, good, neutral, bad, very bad). The pipeline is illustrated in Fig. \ref{fig:pipe}. This approach allows for effective classification even when the number of ideal techniqe examples is small. We demonstrate a binary classification (good, bad) in this work.
\begin{equation}
\text{classification} = \begin{cases} 
      \text{Good} & \text{if } \frac{1}{J} \sum_{j=1}^J \mathbb{1}(z_{i,j} < \tau) > \theta \\
      \text{Bad} & \text{otherwise}
   \end{cases}
\end{equation}
Here $\tau$ is a z-score threshold, $\theta$ is the proportion of joints that must meet the threshold for a good classification, and $\mathbb{1}(z_{i,j} < \tau)$ is the indicator function. 

% \begin{table}
% \centering
% \caption{Accuracy Comparison of Models for Different Attributes}
% \label{tab:comparison}
% \begin{tabular}{lccc}
% \hline
% Model & Stance & Weight Transfer & Shot Quality \\
% \hline
% Llava-v1.6-7b & 0.26 & 0.24 & 0.21 \\
% GPT4-V & 0.64 & 0.38 & 0.23 \\
% \poze (Ours) & \textbf{0.77} & \textbf{0.61} & \textbf{0.74} \\
% \hline
% \end{tabular}
% \end{table}

\begin{figure}
    \centering
    \includegraphics[width=0.85\linewidth]{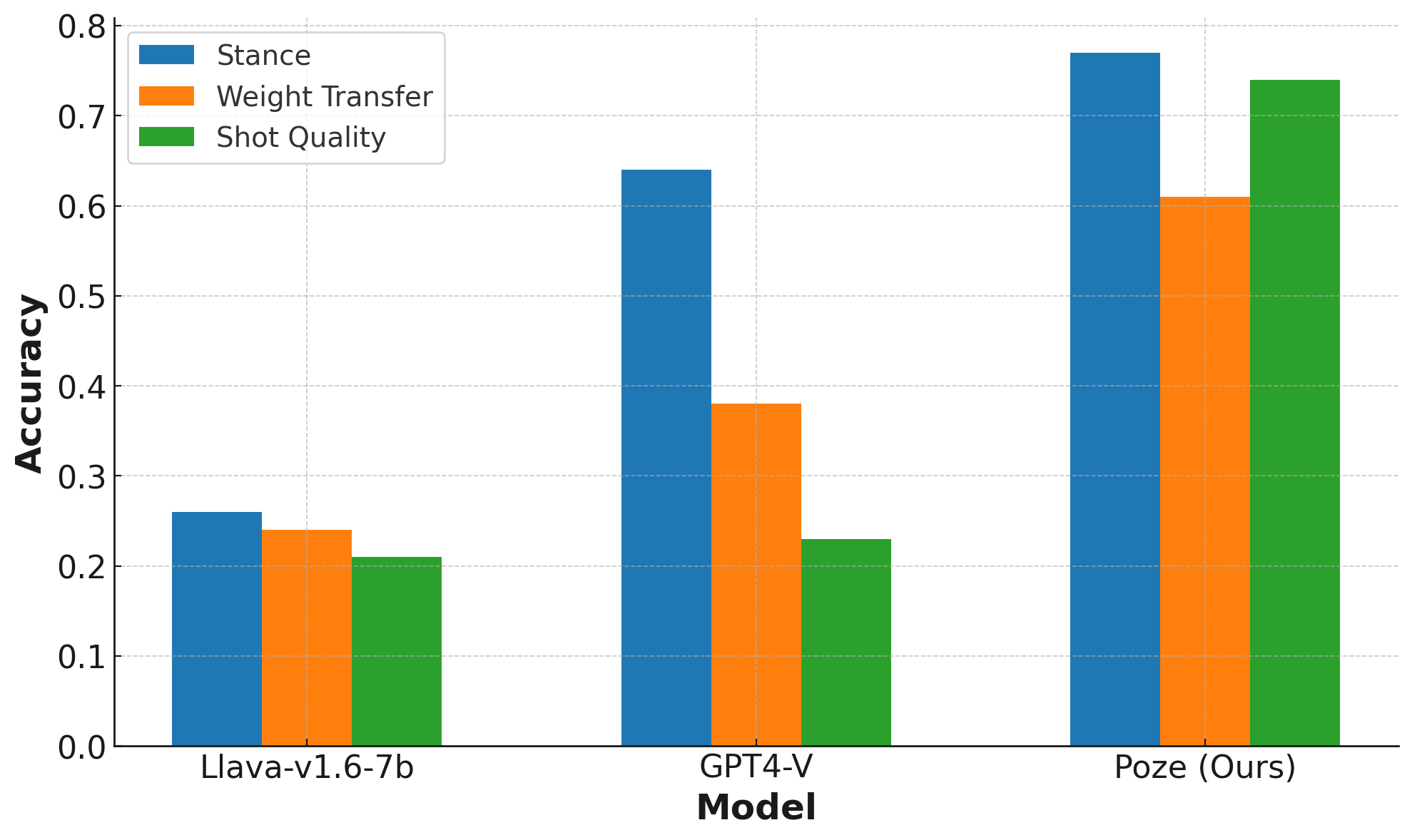}
    \caption{Accuracy comparison for modelled Attributes.}
    \label{fig:results}
\end{figure}

\section{Results and Discussion}
We curated a dataset of 287 labeled cricket shot videos, covering four cricket batting techniques (frontfoot defense, backfoot defense, drive, and miscellaneous). Each technique includes approximately 30 ‘ideal’ technique videos, modeled across four attributes suggested by coaches (stance, weight transfer, shot quality). This collection constitutes the ‘training’ set. For attribute classification, we evaluated \textbf{\poze} against the Visual Question Answering framework IG-VLM \cite{igvlm}, using GPT-4V and LLaVa-v1.6-7b as the large vision model backbones. The results, shown in Fig. \ref{fig:results}, are averaged for class imbalance across the four techniques for each attribute, demonstrating that \poze outperforms the VLMs.

%We curated a dataset of 287 labelled cricket shot videos, spanning four cricket batting techniques (frontfoot defence, backfoot defence, drive and misc.) Each technique has $\sim$30 `ideal' technique videos, modelled across four attributes suggested by coaches (stance, weight transfer, shot quality), and this constituted the `training' set. For attribute classification, we compared \poze against Visual Question Answering framework IG-VLM \cite{igvlm} using GPT4-V and LLaVa-v1.6-7b as the Large Vision Model backbones. The results were class imbalance averaged across the four techniques for each attribute. As shown in Fig. \ref{fig:results}, \poze outperforms the VLMs.

\section{Conclusion and Future Work}

The data-efficient approach of \textbf{\poze} outperforms resource-intensive Video Question Answering (VQA) frameworks, showing promise for democratizing sports technique analysis. Beyond sports, \poze has applications in various motion-based activities, such as dance, yoga, and physiotherapy, making advanced analysis accessible to a broader audience. Future work will explore expanding to other sports and integrating tracking for sports equipment like bats, rackets, and balls.

\begin{acks}
This work was supported in part by the PRAYAS Program under the Cyber-Physical Systems scheme at iHub Drishti Foundation.

\end{acks}
\bibliographystyle{ACM-Reference-Format}
\bibliography{mybib}
\end{document}